\pdfoutput=1

\documentclass[11pt,a4paper]{article}
\usepackage{emnlp2021}

\usepackage{times}
\usepackage[letterspace=-75]{microtype}
\usepackage{booktabs,graphicx,multirow}
\usepackage{old-arrows}

\usepackage{latexsym}

\usepackage{microtype}

\usepackage{color}

\def\ourexample#1{``\emph{#1}''}
\newcommand{\webpage}{\url{https://ad.cs.uni-freiburg.de/publications}}

\usepackage[normalem]{ulem}

\usepackage{datetime}
\usepackage{enumitem}
\usepackage{tabularx}
\usepackage{graphicx}
\usepackage{subcaption}
\usepackage{tikz}
\usepackage{url}
\usepackage{amsmath}
\usepackage{arydshln}  

\newfont{\mycrnotice}{ptmr8t at 7pt}
\newfont{\myconfname}{ptmri8t at 7pt}

\title{Tokenization Repair in the Presence of Spelling Errors\\[-5mm]}

\author{
Hannah Bast, Matthias Hertel, Mostafa M. Mohamed \\
       University of Freiburg \\
       Freiburg, Germany \\
       \texttt{\{bast,hertelm,amin\}@cs.uni-freiburg.de}
}

\date{}

\begin{document}

\maketitle
\thispagestyle{plain}  

\vspace*{-20mm}

\begin{abstract}
We consider the following \emph{tokenization repair problem}:
Given a natural language text with any combination of missing or spurious spaces, correct these.
Spelling errors can be present, but it's not part of the problem to correct them.
For example, given: \ourexample{Tispa per isabout token izaionrep air}, compute \ourexample{Tis paper is about tokenizaion repair}.

We identify three key ingredients of high-quality tokenization repair, all missing from previous work:
deep language models with a bidirectional component, training the models on text with spelling errors, and making use of the space information already present.
Our methods also improve existing spell checkers by
fixing not only more tokenization errors but also more spelling errors:
once it is clear which characters form a word,
it is much easier for them to figure out the correct word.

We provide six benchmarks that cover three use cases
(OCR errors, text extraction from PDF, human errors)
and the cases of partially correct space information and all spaces missing.
We evaluate our methods against the best existing methods and a non-trivial baseline.
We provide full reproducibility under \webpage\ .

%

\end{abstract}

\section{Introduction}\label{sec:intro}
\setlength{\abovedisplayskip}{2mm}
\setlength{\belowdisplayskip}{2mm}

Tokenizing a given text into words is the first step in many natural language processing applications, including: search engines, translation services, spell checkers and all kinds of learning tasks performed on text. 
This tokenization is typically performed by the following simple method or a variant of it:
define a set of word characters and take each maximal sequence of word characters as one token. %
For example, for
\begin{eqnarray}
\textrm{This algoritm runs in linear time}
\end{eqnarray}
a simple such tokenization yields the six words
\begin{eqnarray*}
\textrm{This, algoritm, runs, in, linear, time.}
\end{eqnarray*}
\newpage \vspace*{-20mm}\noindent
Note the spelling error in the second word.
Spelling correction is not part of tokenization.
We come back to this important aspect in Section \ref{sec:intro:spelling}.

Missing and spurious spaces are common errors in digital text documents.
We refer to the union of both types of errors as \textit{tokenization errors}.
Here is a variant of the sentence above with one missing space and one spurious space:
\begin{eqnarray}
\textrm{This algor itm runsin linear time}
\end{eqnarray}
In this paper, we consider the following \emph{tokenization repair} problem:
Given a sequence of characters representing a natural language text, with an arbitrary amount of missing and spurious spaces and possibly also with spelling errors, compute the variant of the text with correct spacing.
For example, given (2) above, compute (1).

Tokenization repair can be considered as a generalization of the \emph{word segmentation} problem, where the text is given without any space information.
Indeed, we also evaluate our methods on this special case in Section \ref{sec:evaluation}.%

\subsection{Sources of tokenization errors}\label{sec:intro:sources}

Tokenization errors are typical in texts that are digitized by Optical Character Recognition (OCR) techniques.
For example, tokenization errors are known to be frequent in the ACL anthology corpus \cite{nastase} and in digitized newspapers \cite{historical, swedish}.
Many OCR error correction methods can not deal with tokenization errors, and it is stated in \citet{helsinki} that:\\[2mm]
%
{\normalsize "A limitation of our approach is that it cannot do word segmentation in case multiple words have been merged together as a result of the OCR process.  However, this problem is complex enough on its own right to deserve an entire publication of its own and is thus not in the scope of our paper."}\\[2mm]
%
Portable document formats like PDF specify the position of the characters on a page.
When extracting text from such formats,
space positions must be inferred from the distance between the characters' bounding boxes,
which is error-prone.
Tokenization errors can also be found in human-typed texts.
The fraction of these errors among misspellings was found to be 15\,\% in \citet{kukich}.

Tokenization errors degrade the performance of any natural language processing (NLP) system, if it does not account for them.
A search engine will not find \ourexample{algorithm} in a document containing \ourexample{algo rithm}.
Syntax parsers and word labelers will not give the correct results if a word is split into multiple tokens, or multiple words merged into one.
A text classifier based on word statistics or word vector representations will fail to retrieve statistics or vector representations for wrongly tokenized words, which can result in wrong classifications.

\subsection{Tokenization and Spelling Correction}\label{sec:intro:spelling}

Existing spelling correction tools either assume correct tokenization or fix tokenization errors only to a limited extent;
see our evaluation in Section \ref{sec:evaluation}.
We show that by using our tokenization repair as a pre-processing,
not only are more tokenization errors fixed (obviously), but also more spelling errors.
The reason is that once it is clear which characters form a word, it is much easier to figure out the correct word.
For example, the best spelling corrector from our evaluation corrects the passage \ourexample{Mqr ymay have k i ssecl John}
from the ACL anthology to \ourexample{Mqr may have k i secl John} without prior tokenization repair
and to the correct \ourexample{Mary may have kissed John} after tokenization repair.

It seems that, ideally, tokenization and spelling errors should be fixed together.
However, this appears to be a very hard problem.
Our own approaches can be adapted to also fix spelling errors,
but only with an impractical large running time.
There is a fundamental reason for this:
because of the very many possible interpretations,
the size of the \emph{beam} (containing the best partial corrections of the sequence) needs to be very large, 
in order not to miss the correct solution; see Section \ref{subsec:character-based}.
All other tools we know of that consider both tokenization and spelling errors (and some even grammar errors),
fare very poorly on passages with tokenization and spelling errors combined.

\subsection{Contributions}

We consider these as our main contributions:

\begin{itemize}[parsep=-1.0mm,leftmargin=0mm,itemindent=4mm,topsep=0pt]


\item We consider the problem of tokenization repair in the presence of spelling errors.
Unlike previous work, our model can make use of the space information already present.
We can also solve the classical word segmentation problem (all spaces removed) better.
We show that it is crucial to train on text with (the right kind and dose of) spelling errors.

\item In previous work, forward models combined with a beam search gave the best results.
We present an elegant idea to realize a bidirectional model; see Figure \ref{fig:bidirectional-model} for an illustration.
This is tricky for a task like tokenization repair, which involves changing the sequence while correcting it. 


\item We provide six benchmarks that cover all our use cases
(OCR errors, text extraction from PDFs, human errors)
and different degrees of available space information
(partially correct or all spaces missing).
We make use of existing benchmarks wherever possible (manually augmenting some by a ground truth)
and create new benchmarks with realistic error models.

\item We compare our approach with the best existing methods
(from the literature) and tools (commercial and open-source).
We evaluate both the quality of the tokenization repair and
how existing spelling correctors perform much better
when the space errors are fixed first.

\item We provide a single model, trained across multiple corpora,
that produces good results across all benchmarks
without the need for any fine-tuning or hyperparameter optimization.

\item Our code, data, benchmarks and trained models are available under \webpage.
It includes a Docker setup that allows an easy replication of all our results,
a web application that allows an interactive error analysis for all methods and benchmarks,
and a version of the ACL anthology corpus where the tokenization errors were corrected by our best method.

\end{itemize}

\section{Related Work}\label{sec:related}
A beam search with neural and character $n$-gram language models is used for word segmentation in \citet{doval}.
In Section \ref{sec:evaluation}, we evaluate our own implementation of this approach on our benchmark
and we also compare against their results on their benchmark.
We improve on their approach in several respects: integrating a bidirectional model (which is not trivial),
considering the given spaces in the input (they remove all spaces), and explicitly addressing typos
by incorporating error models in training (they test their approach on tweets, but do not explicitly handle typos).

Tokenization repair on the ACL anthology corpus is done in \citet{nastase}
with a neural machine translation model translating from the sequence without spaces to the sequence with spaces.
They also remove all spaces from the input text, thus discarding valuable information.
In Section \ref{sec:evaluations:results}, we compare our results against theirs.

A beam search with a word bigram language model, instead of a character-based language model, is used in \citet{croatian} to correct missing spaces in Croatian texts that were digitized by OCR.
In \citet{historical}, $n$-gram statistics are used to determine when to split an out-of-vocabulary token into two words.
By using neural language models, we extend the scope of this context beyond the boundaries of $n$-gram models.

Recent work on Chinese and Arabic word segmentation uses bidirectional neural network models to predict word boundaries, e.g. based on bi-LSTMs \cite{chinese-lstm, arabic-lstm} or a pre-trained BERT model \cite{chinese-bert}.
These models are not directly applicable for our task of tokenization repair for the English language,
since our inputs contain spurious and missing spaces
and the English BERT model \cite{bert} is pre-trained on text with correct tokenization.

There is a large body of research on OCR postcorrection \cite{ocr-tong, ocr-spell, ocr-niklas, ocr-kissos, icdar2017, ocr-dong, icdar2019, helsinki, ocr-nguyen19, ocr-nguyen}, which usually does not address tokenization errors explicitly. Only few publications provide code so that we can evaluate their tokenization repair capabilities.

\section{Approach}\label{sec:approach}

\newcommand{\ins}{\mathit{ins}}
\newcommand{\del}{\mathit{del}}
\newcommand{\msp}{\text{\textvisiblespace}}
\newcommand{\bi}{\mathit{bi}}
\newcommand{\pbi}[1]{\overleftrightarrow{p_{#1}}}
\newcommand{\pf}{\overrightarrow{p}}

Our approach is a beam search based on deep character-based models, unidirectional and bidirectional.

\subsection{Character-based models}
\label{subsec:character-based} 
We represent the strings as sequences of one-hot encoded characters, where we use
the 200 most frequent characters, while replacing the others by a special character UNK
for unknown characters. Sequences are appended with start and end of sentence
special characters (SOS and EOS).

\subsubsection{Unidirectional language models}
Character-based language models estimate the probability of a string to occur
in some language based on the probabilities of the individual characters in the string.
Following \citet{graves:generating}, we implement these models
as recurrent neural networks, using LSTM cells.
Our architecture consists of an LSTM cell with 1024 units,
followed by a dense layer (with 1024 units and ReLU activation)
and a softmax output layer for character classification.
This architecture has 6,287,563 trainable parameters, which are
trained with the categorical cross entropy loss.
The model predicts $\pf(s | c)$, which is the probability that a character $s$ comes
after a context $c$. Moreover, we define $\pf(\msp s | c) := \pf(s | c \msp) \cdot \pf( \msp | c)$
as the probability that a space and character $s$ come after context $c$.

\subsubsection{Bidirectional sequence labeling model} \label{subsec:bidirectional}

We utilize a bidirectional model that predicts the probability $\pbi{i}$ of having a space before the $i^\text{th}$ character when the whole sequence of non-space characters is the input. Our architecture consists of a bidirectional LSTM
cell with 1024 units, followed by a dense layer (with 1024 units and ReLU activation) and a sigmoid output unit.
This architecture has 12,158,980 trainable parameters,
which are trained with the binary cross entropy loss.



\subsection{Beam search} \label{subsec:beam-search}
Beam search is a search algorithm similar to breadth-first search, but instead of
maintaining all search states at a given level, it maintains only the best $b$ states,
which correspond to an estimation of the best $b$ partial solutions 
(called \emph{beams}) \cite{beam-search}.
We introduce two variants of beam search, unidirectional (UNI) and bidirectional (BID).

\paragraph{Correction procedure:}
Given a mistokenized string $Q$, with its corresponding sequence of $m$ non-space characters $T$,
the procedure executes beam search for $m$ levels. 
At level $i$, given a partial solution's search state $(S_{i-1}, R_{i-1})$ of accumulated
score and partial solution string, two candidate extensions are created:
\begin{enumerate}[wide, labelwidth=0pt, labelindent=0pt, noitemsep, topsep=0pt]
\item Adding $T_i$ without space, which results in:
\begin{center}
    \vspace{-0.2cm}
  $S_i = S_{i - 1} - \log \pf(T_i | R_{i-1}) + P_\del$
    \\
    $R_i = R_{i - 1}  T_i$
    \vspace{-0.1cm}
\end{center}
\item Adding a space before $T_i$, which results in:
\begin{center}
\vspace{-\parskip}
    \vspace{-0.2cm}
    $S_i' = S_{i-1} - \log \pf(\msp T_i| R_{i-1}) + P_\ins$
    \\
    $R_i' = R_{i - 1} \msp \text{ } T_i$
    \vspace{-0.1cm}
\end{center}
\end{enumerate}
$P_\del$ and $P_\ins$ are non-negative penalties that are
used only when the introduced extension is not originally in $Q$,
otherwise they are equal to $0$.
In other words, considering the character $Q_{j-1}$ preceding $T_i$ in the input sequence, $P_\ins$ is used when
$Q_{j-1} \neq \msp$ and $P_\del$ is used when $Q_{j - 1} = \msp$.
The penalties regularize the effect of making too many edits.
These equations correspond to UNI.
\\
\\
The final solution $R^{*}$ 
is the estimated corrected sequence of lowest penalized negative log-likelihood score
(highest probability):
\begin{center}
    $-\log p(R^{*}) + n_\ins P_\ins + n_\del P_\del$
\end{center}
where $n_\ins$ is the number of space insertions and $n_\del$ is the number of
space deletions, and
$$p(R^{*}) = \prod_{i=1}^{|R^{*}|} \pf(R^{*}_i | R^{*}_{1:(i-1)})$$
The time complexity is $\mathcal{O}(|Q|\cdot b)$, because we process $2b$ candidates
at $m$ levels ($m \leq |Q|$).
We use a beam size $b=5$ in our implementation. As a result, the algorithm runs in linear time.

\paragraph{Beam search bidirectional (BID):} 
This method combines UNI with the bidirectional labeling model introduced in section \ref{subsec:bidirectional}.
We process the non-space characters $T$ using the bidirectional model, then we modify the beam search formulas: 

\begin{center}
    $S_i = S_{i - 1} - \log (\pf(T_i | R_{i-1}) \cdot (1 - \pbi{i})) + P_\del$
\\
    $S_i' = S_{i-1} - \log (\pf(\msp T_i| R_{i-1}) \cdot \pbi{i}) + P_\ins$
\end{center}

\noindent As stated earlier, the penalties are only used if they correspond to changes that are not originally in the given text.
Figure \ref{fig:bidirectional-model} shows an example illustrating these equations.

\begin{figure}
    \small
 \hspace{-0.2cm}
\begin{tikzpicture}[node distance=1cm]
 \tikzset{char/.style = {draw, circle, scale=0.68, minimum size=0.7	cm}}
 \node (input) {$Q$};
 \node[char] (h) [right of=input, node distance=1cm] {h};
 \node[char] (e) [right of=h] {e};
 \node[char] (l1) [right of=e] {l};
 \node[char, inner sep=5pt] (space) [right of=l1] {\msp};
 \node[char] (l2) [right of=space] {l};
 \node[char, inner sep=4pt] (o) [right of=l2] {o};
 \node[char] (w) [right of=o] {w};
 \node[char] (space2) [right of=w] {\msp};
 \node[char, inner sep=4pt] (o2) [right of=space2] {o};
 \node[char] (r) [right of=o2] {r};
 \node[char] (d) [right of=r] {d};

 \node (fwd) [below of=input] {$R$};
 \node[char, inner sep=0pt] (fh) [right of=fwd, node distance=1cm] {$R_1$};
 \node[char, inner sep=0pt] (fe) [right of=fh] {$R_2$};
 \node[char, inner sep=0pt] (fl1) [right of=fe] {$R_3$};
 \node[char, inner sep=0pt] (fl2) [right of=fl1, node distance=2cm] {$R_4$};
 \node[char, inner sep=0pt] (fo) [right of=fl2] {$R_5$};
 
 \node (bid) [above of=input] {$\pbi{}$};
 \node[char, inner sep=0pt] (bh) [right of=bid] {$\pbi{1}$};
 \node[char, inner sep=0pt] (be) [right of=bh] {$\pbi{2}$};
 \node[char, inner sep=0pt] (bl1) [right of=be] {$\pbi{3}$};
 \node[char, inner sep=0pt] (bl2) [right of=bl1, node distance=2cm] {$\pbi{4}$};
 \node[char, inner sep=0pt] (bo) [right of=bl2] {$\pbi{5}$};
 \node[char, inner sep=0pt] (bw) [right of=bo] {$\pbi{6}$};
 \node[char, inner sep=0pt] (bo2) [right of=bw, node distance=2cm] {$\pbi{7}$};
 \node[char, inner sep=0pt] (br) [right of=bo2] {$\pbi{8}$};
 \node[char, inner sep=0pt] (bd) [right of=br] {$\pbi{9}$};

 \node (formula1) [below of=fo, node distance=0.5cm, yshift=-0.2cm] {$S_6 = S_5 - \log (\pf(\text{w}|\text{hello}) \cdot (1-\pbi{6}))$};
 \node (formula2) [below of=formula1,node distance=0.5cm] {$S_6' = S_5 - \log (\pf(\msp \text{w}|\text{hello}) \cdot \pbi{6}) + P_{\text{ins}}$};
 \node (formula3) [below of=formula2, node distance=0.6cm] {$S_4 = S_3 - \log (\pf(\text{l}|\text{hel}) \cdot (1-\pbi{4})) + P_{\text{del}}$};
 \node (formula4) [below of=formula3,node distance=0.5cm] {$S_4' = S_3 - \log (\pf(\msp \text{l}|\text{hel}) \cdot \pbi{4})$};

 \draw[->] (fh) -- (fe);
 \draw[->] (fe) -- (fl1);
 \draw[->] (fl1) -- (fl2);
 \draw[->] (fl2) -- (fo);

 \draw[<->] (bh) -- (be);
 \draw[<->] (be) -- (bl1);
 \draw[<->] (bl1) -- (bl2);
 \draw[<->] (bl2) -- (bo);
 \draw[<->] (bo) -- (bw);
 \draw[<->] (bw) -- (bo2);
 \draw[<->] (bo2) -- (br);
 \draw[<->] (br) -- (bd);

 \draw[->] (h) -- (fh);
 \draw[->] (e) -- (fe);
 \draw[->] (l1) -- (fl1);
 \draw[->] (l2) -- (fl2);
 \draw[->] (o) -- (fo);

 \draw[->] (h) -- (bh);
 \draw[->] (e) -- (be);
 \draw[->] (l1) -- (bl1);
 \draw[->] (l2) -- (bl2);
 \draw[->] (o) -- (bo);
 \draw[->] (w) -- (bw);
 \draw[->] (o2) -- (bo2);
 \draw[->] (r) -- (br);
 \draw[->] (d) -- (bd);

\end{tikzpicture}
\vspace{-0.5cm}
\caption{
A snapshot of one beam fixing the string with typo ``hel low ord'', as explored by bidirectional beam search.
For the beam $R_5=$\ourexample{hello}, the equations for $S_6, S_6'$ show the
updated scores for the two new states of not inserting a space and inserting a space, respectively.
For the earlier beam $R_3=$\ourexample{hel}, the equations for $S_4, S_4'$ show
the updated scores for deleting the space and keeping it, respectively.
}
\label{fig:bidirectional-model}
\vspace{-0.1cm}
\end{figure}

\paragraph{Penalty optimization:}
The penalties $P_\ins$ and $P_\del$ are optimized using a development set of mistokenized sequences and their ground truth.
We simulate a beam search assuming that the left context is always predicted correctly,
and that the procedure takes a decision after processing the next two characters.
Given the ground truth string $Q$, for every non-space character $Q_i$ and its previous non-space character $Q_j$,
the space probability $p_s$ and non-space probability $p_n$ are computed:
\begin{center}
    $p_s = \pf(\msp Q_i|Q_{1:j}) \cdot \pf( Q_{i+1} | Q_{1:j} \msp Q_i ) \cdot \pbi{i'}$
    $p_n=\pf(Q_i|Q_{1:j}) \cdot \pf( Q_{i+1} | Q_{1:j} Q_i ) \cdot (1 - \pbi{i'})$
\end{center}
where $i'$ is the position of the non-space character corresponding to $Q_i$.
These equations are for BID; the last term in both equations is excluded for UNI.
\\
If the space is present in the input sequence, the scores $S$ and $S'$ of the candidate sequences without and with a space are:
\begin{center}
    \vspace{-0.0cm}
 $S = - \log p_n + P_\del$ \\
 $S' = - \log p_s$
    \vspace{-0.0cm}
\end{center}
The space gets deleted if $P_\del < \log p_n - \log p_s$.
If the space is not present in the input, then: 
\begin{center}
    \vspace{-0.0cm}
 $S = - \log p_n$ \\
 $S' = - \log p_s + P_\ins$
    \vspace{-0.0cm}
\end{center}
The space gets inserted if $P_\ins < \log p_s - \log p_n$.
%
%

We perform a grid search on $P_\ins$ and $P_\del$
in the range $[0, 20]$ with step size $0.1$,
and take the combination that maximizes the sequence accuracy
defined in Section \ref{sec:evaluation:metrics:tokenization}.
We optimize sequence accuracy instead of F-score,
because that gives better results on benchmarks with very few errors,
where the model must be very conservative.

\subsection{Baseline approaches}\label{sec:baselines}

We evaluate the following three baselines.
We also tested a greedy algorithm but omit the results because it consistently performed much worse than the dynamic-programming baseline.

\paragraph{Dynamic programming bigram model:}
This baseline uses the Viterbi algorithm \cite{viterbi} with a word bigram model.
First, all possible words (substrings of length $\leq 20$ with non-zero unigram frequency in the training data) are located in the sequence without spaces.
The states of the Viterbi algorithm are equivalent to the words.
A transition between two states is possible if the next word starts at the end of the first word.
State transition probabilities are determined by a combination of a unigram and a bigram model, with probabilities $p_\mathit{uni}$ and $p_\mathit{bi}$ estimated on Wikipedia:
\begin{center}
    $p(w_{i+1}|w_i) = \frac{1}{2} (p_\mathit{uni}(w_{i+1}) + p_\mathit{bi}(w_{i+1}|w_i))$
\end{center}
The output is the most likely segmentation of the sequence into words.

\paragraph{Wordsegment:} 
\textit{Wordsegment}\footnote{\url{https://pypi.org/project/wordsegment/}} is an open-source library, based on \citet{beautifuldata}, that
uses unigram and bigram frequencies to segment words.

\paragraph{Google:}
To compete with a commercial spell checker, we copy the erroneous sentences into a Google document\footnote{\url{https://docs.google.com}, access. Sept. 8, 2021} and accept all suggested edits.
We also evaluated the widely used tools Hunspell, TextRazor, and Grammarly, as well as the OCR post-correction system Natas \cite{helsinki}.
However, for our datasets, Google yielded the best corrections (for both tokenization and spelling errors).

\section{Datasets and training}\label{sec:datasets}
\pdfoutput=1

\def\ptoko{p_{\text{space}}}
\def\ptypo{p_{\text{spell}}}

\subsection{Datasets}\label{sec:evaluation:dataset}

\paragraph{ACL anthology corpus:}
The ACL corpus is extracted from scientific articles published between 1965 and 2012 \cite{acl-corpus}.
The publications were scanned and parsed using OCR, and hence have many typical OCR errors.
We manually corrected the tokenization and spelling of 500 sequences for development and penalty optimization, and 500 sequences as a test set.

\paragraph{arXiv:}
We used the benchmark generator from \citet{PDFs} to generate
the text from 910,000 articles from arXiv
(parsed from \LaTeX\ files and serving as our ground truth)
as well as the text extracted from the corresponding PDF files
(using \emph{pdftotext} from \citet{pdftotext}).
The files were split into paragraphs, which were matched with the ground truth
by searching for a text span that differed from the paragraph only by spaces.
We thus obtain 64,965,651 sequences for training,
and 10,000 sequences each for penalty optimization, development and test.

\paragraph{Wikipedia:}
We extracted the articles from Wikipedia\footnote{\url{https://dumps.wikimedia.org/enwiki}, dump of June 20, 2019}
using WikiExtractor \cite{attardi:wiki-extractor},
and split them into sentences using NLTK \cite{nltk}.
We did a pre-processing to remove sentences that were incomplete or contained markup.
We thus obtain 43,103,197 sequences for training, and 10,000 sequences each for penalty optimization, development and test.

\medskip
We verify the quality of the generated ground truths for Wikipedia and arXiv on 100 sequences from the development sets.
Both ground truths have few errors:
The 100 sequences from Wikipedia contain no tokenization errors and two spelling errors. The 100 sequences from the arXiv ground truth contain three tokenization errors and no spelling errors.

\subsection{Error injection}\label{sec:evaluation:noise}

We inject different types of errors into clean text for two purposes:
(1) To create noisy training data.
(2) To create large synthetic benchmarks with OCR errors and spelling errors (see Section \ref{sec:evaluation:benchmarks}).


\paragraph{OCR errors:}

We consider three kinds of OCR errors:
tokenization errors, hyphenation errors (a spurious hyphen due to hyphenation at the end of a line), and character replacement errors (insertions, replacements or deletions of one or more characters, where no spaces are involved).

We estimate the probabilities for these errors on our ACL development set.
For hyphenation errors, we divide the total number of spurious hyphens by the number of hyphenable tokens. For space insertions, space deletions, and character replacements, we compute the error rates (number of errors divided by the number of characters) per span of $l$ continuous tokens.

We derive the rules for character replacements and their relative frequency by a comparison of the ACL corpus with a cleaned version\footnote{\url{https://web.eecs.umich.edu/~lahiri/acl_arc.html}, we found the corrections to be incomplete, and therefore do not use the cleaned corpus for training or evaluation, but to derive statistics about typical OCR errors.},
and from the ICDAR 2017 and ICDAR 2019 OCR post-correction benchmarks \cite{icdar2017, icdar2019}.

To inject errors, we first hyphenate each token with the estimated probability of 3.5\,\% using \textit{PyHyphen}.\footnote{\url{https://pypi.org/project/PyHyphen/}}
We pick each token as the beginning of an erroneous span with the estimated probability of 5.8\,\%.
We sample the span length $l$, space insertion, space deletion and character replacement rates from the collected data,
and introduce errors into the next $l$ tokens accordingly. Character replacements are sampled following their frequency in the ACL and ICDAR datasets.
No tokenization errors are injected for the training data.



%

\paragraph{Human errors:}

We use another model to inject human errors. Each word gets replaced by a misspelling from a typo collection with 10\,\% probability. The collection contains 228,414 spelling errors from Peter Norvig\footnote{\url{https://norvig.com/ngrams/spell-errors.txt}}, Twitter\footnote{\url{http://luululu.com/tweet/typo-corpus-r1.txt}} and \citet{github-corpus}, which we split equally into a training/development set and a test set.
For the training data, we also generate random spelling errors (character insertions, deletions, replacements or swaps).
For the synthetic benchmarks, we inject on average $p_t$ tokenization errors per token, with a constant $p_t$.

\subsection{Models training}\label{sec:evaluation:models}

We use the two beam search approaches described in Section~\ref{subsec:beam-search}: 
a unidirectional model (UNI), and the unidirectional model combined with the bidirectional sequence labeling model
described in Section~\ref{subsec:character-based} (we call this combination BID).
For each of these, we train two variants:
one using a combination of the clean \emph{arXiv ground truth} (from the \LaTeX\ source files) and \emph{Wikipedia} datasets from Section \ref{sec:evaluation:dataset},
and the other using the same datasets, but with OCR and spelling error noise injected as described in the previous subsection.

For each approach, we optimize the penalties $P_\ins$ and $P_\del$
on the penalty optimization set for every benchmark with spaces; see Section \ref{sec:evaluation:dataset}.
Additionally, we consider a model with fixed (benchmark-independent) penalities,
averaged across the five benchmarks with spaces.
When there are no spaces (which is trivial to detect),
we set $P_\ins = P_\del = 0$.
In summary:
\\[1mm]
\begin{tabular}{ll}
UNI            & UNI trained on clean text\\
UNI+           & UNI trained on noisy text\\
BID            & BID trained on clean text\\
BID+           & BID trained on noisy text\\
BID+ The One   & BID+ with fixed penalties\\
\end{tabular}\\[1mm]
The models were trained for one epoch,
which took 86 hours for the unidirectional and 144 hours for the bidirectional models
on a NVIDIA Titan X GPU.
The training was performed using the Adam optimization algorithm \cite{adam}, 
with learning rate 0.001, and mini-batch size 128.
The sequences were cut after 256 characters, while shorter sequences were padded with EOS symbols that got masked in the loss function.
The models are implemented using TensorFlow \cite{tensorflow}.
The unidirectional language model has $67.7\,\%$ character accuracy,
$88.8\,\%$ top-5 character accuracy and $1.099$ categorical cross-entropy.

\section{Evaluation}\label{sec:evaluation}
\subsection{Benchmarks}\label{sec:evaluation:benchmarks}

We evaluate our methods on six benchmarks with different kinds 
of spelling and space errors.
The datasets and ground truths behind these benchmarks,
as well as the division into test, development, and training set,
are described in Section \ref{sec:evaluation:dataset}.
We here describe the corrupt sequences used as inputs to our models.

\paragraph{ACL:}
The 500 test sequences from the ACL anthology dataset are considered.
This benchmark contains many tokenization and OCR errors,
often in combinations that are very hard to fix.

\paragraph{arXiv OCR:}
The 10,000 ground truth sequences from the \emph{arXiv} test set are considered.
We create corrupt sequences by injecting OCR errors into the ground truth
according to our noise model described in Section \ref{sec:evaluation:noise}
(this includes tokenization errors).
We created this benchmark in order to have a larger benchmark than ACL, but with similar properties.

\paragraph{arXiv pdftotext:}
The same 10,000 ground truth sequences are considered, but corrupt sequences taken from
the output of \emph{pdftotext} on the corresponding PDFs.
This benchmark has no spelling errors and few tokenization errors,
so it is hard to make the necessary few corrections, yet avoid false positives.

\paragraph{Wikipedia (three variants):}
For the corrupt sequences of \textit{Wiki+} and \textit{Wiki+ no\msp},
we inject spelling errors into the Wikipedia ground truth using the human error model described in \ref{sec:evaluation:noise}.
For \textit{Wiki+}, we inject tokenization errors with a realistic (see Section \ref{sec:intro:sources}) rate of $p_t=0.01$ per token.
For \textit{Wiki+ no\msp}, we remove all spaces,
which is the scenario from previous work and the \emph{word segmentation} problem.
\textit{Wiki} has no spelling errors and a tokenization error rate of $p_t=0.1$, similar as in the PDF extraction scenario.

\smallskip
Statistics of the tokenization errors in the six benchmarks are given in Table \ref{tab:tokenization-benchmarks}.
We have also evaluated our approaches on the ICDAR benchmarks (Section \ref{sec:evaluation:noise}),
but do not report them due to severe issues with the ground truth.%
\footnote{Many corrections were missing and the word order was often significantly changed.
As a relatively minor complication, some of the text uses old language and symbols not encountered in our training data.}

\newcommand{\h}[1]{\multicolumn{1}{c|}{#1}}

\begin{table}
 \centering
 \small
 \setlength{\tabcolsep}{4pt}
 \renewcommand{\arraystretch}{1.1}
 \begin{tabular}{|l|r|r|r|r|}
  \hline
   Benchmark                    & \h{Sequences} & \h{Erroneous} & \h{Spurious} & \h{Missing}  \\
  \hline
   ACL                          &           500 &     190 &  1,160 &     297 \\
   arXiv OCR                    &        10,000 &   3,570 & 14,242 &   3,798 \\
   arXiv \scriptsize{pdftotext} &        10,000 &   1,274 &  2,355 &     594 \\
   Wiki                         &        10,000 &   6,502 &  7,681 &   7,261 \\
   Wiki+                        &        10,000 &   1,310 &    726 &     727 \\
   Wiki+ no \msp                &        10,000 &   9,590 &      0 & 141,750 \\
  \hline
 \end{tabular}
 \caption{Statistics of the tokenization repair benchmarks. Erroneous = sequences with tokenization errors, Spurious = spurious spaces, Missing = missing spaces.}
 \label{tab:tokenization-benchmarks}
\end{table}

\begin{table}
 \centering
 \small
 \setlength{\tabcolsep}{4pt}
 \renewcommand{\arraystretch}{1.1}
 \begin{tabular}{|l|r|r|r|r|r|}
  \hline
   Benchmark                    & \h{Sequences} & \h{Tokens} & \h{TE} & \h{SE} & \h{ME} \\
  \hline
   ACL                          &           500 &     12,990 &    914 &    451 &    143 \\
   arXiv OCR                    &         1,000 &     22,446 &  1,073 &    571 &    133 \\
   Wiki+                        &         1,000 &     15,369 &    215 &    901 &     19 \\
   Wiki+ no \msp                &         1,000 &     15,369 & 14,192 &      0 &  1,133 \\
  \hline
 \end{tabular}
 \caption{Statistics of the spelling benchmarks. Tokens = number of ground truth tokens, TE = tokenization errors, SE = spelling errors, ME = mixed errors.}
 \label{tab:spelling-benchmarks}
 \vspace{-0.4cm}
\end{table}

\subsection{Metrics}\label{sec:evaluations:metrics}

\subsubsection{Tokenization repair}\label{sec:evaluation:metrics:tokenization}

Given a corrupt input text $C$, a ground truth text $T$ and a predicted text $P$,
a tokenization repair algorithm predicts a set
of space insertions and deletions that ideally would transform $C$ into $T$.
We use two metrics for the evaluation: F-score and sequence accuracy.

\paragraph{F-score:}\label{sec:evaluation:fscore}

We define $\mathrm{edits}(A, B)$ as the space insertions and deletions that transform $A$ into $B$.
If we let $\mathcal{C} = \mathrm{edits}(C, T)$ be the ground truth edit operations
and $\mathcal{P} = \mathrm{edits}(C, P)$ the predicted edit operations,
the number of true positives is $\mathrm{TP} = |\mathcal{C} \cap \mathcal{P}|$,
the number of false positives is $\mathrm{FP} = |\mathcal{P} \setminus \mathcal{C}|$
and the number of false negatives is $\mathrm{FN} = |\mathcal{C} \setminus \mathcal{P}|$.
The F-score is computed as:
$$F(T, C, P) = \frac{2 \cdot \mathrm{TP}}{2 \cdot \mathrm{TP} + \mathrm{FP} + \mathrm{FN}}$$

\paragraph{Sequence accuracy:}
The sequence accuracy is the fraction of sequences that are corrected completely ($P = T$).

\subsubsection{Spelling correction}

We use a word correction F-score as a metric for spelling correction.
We compute the longest common token subsequence of $C$ with $T$ (that is, $C \cap T$), and $P$ with $T$ (that is, $P \cap T$), and define erroneous tokens $E = T \setminus C$, error-free tokens $F = C \cap T$, and correctly predicted tokens $S = P \cap T$.
Then $\mathrm{TP} = |E \cap S|$,
$\mathrm{FP} = |F \setminus S|$
and $\mathrm{FN} = |E \setminus S|$,
and the F-score is computed as above.

We classify each token in $E$ as \textit{tokenization error} if its correction involves only space edits, \textit{spelling error} if only non-space edits, and \textit{mixed} otherwise.
See Table \ref{tab:spelling-benchmarks} for the distribution of the three error types in the benchmarks with spelling errors.

\def\field#1#2{\multicolumn{1}{#1}{#2}}  
\def\nan#1{\field{c#1}{-}}               
\def\b#1{\textbf{#1}}
\def\sup#1{\textsuperscript{#1}}
\def\pz{\phantom{0}}

\begin{table*}[tb]
 \setlength{\tabcolsep}{5.6pt}
 \renewcommand{\baselinestretch}{1.2}\small

 \begin{center}
 \def\pdftotext{\raisebox{1mm}[-1mm]{\scriptsize pdftotext}}
 ~\\[3mm]
 \begin{tabular}{|l|cccccc||cccccc|}
  \hline
     & \multicolumn{6}{c||}{F-score} & \multicolumn{6}{c|}{Sequence accuracy} \\
  \hline
                  &  \multirow{2}{*}{ACL} & arXiv & arXiv      & \multirow{2}{*}{Wiki} & \multirow{2}{*}{Wiki+} & Wiki+
                  &  \multirow{2}{*}{ACL} & arXiv & arXiv      & \multirow{2}{*}{Wiki} & \multirow{2}{*}{Wiki+} & Wiki+ \\
                  &                       &  OCR  & \pdftotext &                       &                        & no $\msp$
                  &                       &  OCR  & \pdftotext &                       &                        & no $\msp$ \\
     \hline
     Do nothing   &  \pz0.0 &  \pz0.0 &  \pz0.0 &  \pz0.0 &  \pz0.0 &  \pz0.0 &    62.0 &    64.3 &    87.3 &    35.0 &    86.9 &  \pz4.1 \\
     Dyn.~Progr.  &    57.4 &    62.2 &    27.5 &    92.6 &    33.0 &    98.0 &    40.2 &    52.9 &    68.1 &    86.2 &    68.6 &    68.6 \\  \hline
     Wordsegment  &    55.2 &    58.2 &    18.7 &    59.1 &  \pz9.9 &    91.1 &    32.2 &    52.0 &    58.3 &    41.1 &    32.0 &    32.0 \\
     Google       &    54.7 &    60.8 &  \pz2.5 &    74.0 &    66.3 &    12.4 &    67.8 &    75.8 &    86.0 &    58.4 &    90.6 &  \pz9.1 \\  \hline
     UNI          &    83.6 &    90.6 &    76.5 &    98.2 &    85.8 &    98.3 &    70.8 &    84.5 &    93.2 &    95.2 &    96.0 &    72.5 \\ 
     UNI+         &    86.0 &    96.1 &    73.5 &    98.3 &    92.7 &    99.1 &    77.6 &    92.8 &    93.4 &    95.7 &    98.0 &    85.3 \\  \hline
     BID          &    88.7 &    94.0 &\bf{85.1}&\bf{99.0}&    86.5 &    98.9 &    76.0 &    87.7 &\bf{94.8}&\bf{97.3}&    96.2 &    80.2 \\
     BID+         &    89.6 &\bf{97.5}&    84.6 &    98.7 &\bf{93.7}&\bf{99.4}&\bf{79.8}&    94.1 &    94.2 &    96.8 &\bf{98.3}&\bf{89.0}\\  \hline
     BID+ The One &\bf{90.6}&\bf{97.5}&    81.8 &    98.9 &    91.5 &\bf{99.4}&    79.6 &\bf{94.2}&    94.1 &    97.2 &    97.7 &\bf{89.0}\\
   \hline
 \end{tabular}
 \renewcommand{\baselinestretch}{1}
 \caption{Micro-averaged F-scores and sequence accuracies in percent for all models (baselines, existing tools, unidirectional, bidirectional) and the six benchmarks from Section \ref{sec:evaluation:benchmarks}.
     The best results for each benchmark are shown in bold.
     All sequence accuracy differences larger than 0.5\,\% in the table are statistically strongly significant ($p < 0.01$ with a paired two-sided randomization test).
     Google's tokenization repair was evaluated on 500 sequences from ACL and 1,000 sequences for all other benchmarks.
     }
 \label{tab:eval-results}
 \vspace{-0.5cm}
 \end{center}
\end{table*}

\subsection{Main results}\label{sec:evaluations:results}

\def\sKB{\,\mathrm{sec\hspace{-0.5mm}/\hspace{-0.5mm}KB}}
Table \ref{tab:eval-results} provides F-scores and sequence accuracies for all our methods on all benchmarks.
We use a beam size of $b=5$ for all beam search approaches; increasing this to $b = 10$ has shown only minimal improvements while doubling the running time.
The average running time of our tokenization repair, measured on an NVIDIA Titan X GPU, is $2.0 \sKB$ for UNI and $2.6 \sKB$ for BID.

The main takeaway from Table \ref{tab:eval-results} is that the bidirectional models beat all four baselines by a wide margin on all benchmarks.
They are also better than their unidirectional counterparts trained on the same data. 
We remark that it is not obvious that the bidirectional methods are the best,
which might be the reason why previous work used unidirectional methods.
A unidirectional model has the advantage that the tokenization errors are incrementally fixed from left to right, so that the language model predictions can be based on text that is (almost) free from such errors.
However, the text after the current position has not yet been repaired, so that predictions from the other direction are based on text with tokenization errors.
Using these predictions actually \emph{deteriorates} the quality of the unidirectional methods.
Our trick was to combine a unidirectional model that makes use of the space information with a bidirectional model that disregards all space information and thus does not have the aforementioned problem.

The other important takeaways from Table \ref{tab:eval-results} are as follows.
The results for previous work are discussed in Section \ref{sec:evaluation:previous}.

\vspace{1mm}
\begin{itemize}[parsep=-0.5mm,leftmargin=0mm,itemindent=4mm,topsep=0pt]

\item Tokenization repair is harder when there are \emph{spelling} errors.
This is especially pronounced for the ACL benchmark, which has many passages so deformed by OCR errors
that there is simply not enough information to reconstruct the correct spacing; 
see Section \ref{sec:evaluation:analysis}.
When there are spelling errors, training on text with spelling errors (UNI+ and BID+)
is crucial, as it enhances the results significantly.

\item Tokenization is also harder when there are more \emph{space} errors,
yet previous work chose to remove all spaces from the text.
Removing all spaces simplifies the approaches (one only has to predict space insertions, no space deletions),
but the price is much worse results when there were actually only few tokenization errors;
see the sequence accuracies of \textit{Wiki+ no \msp} vs. \textit{Wiki+}.

\item The \textit{arXiv pdftotext} benchmark is hard, because $87.3\,\%$ of the sequences have no errors
and it is hard to correct the other $12.7\,\%$ without inserting mistakes into the error-free sequences.
Indeed, Google is very conservative on this benchmark and suggests almost no corrections (hence the very low F-score). 
Wordsegment is bolder, with a result that is much worse than doing nothing.
UNI and BID improve very significantly over doing nothing.

\item The results for ``BID+ The One'' are close to the results of the best models with penalties tailored to the respective benchmark.
This shows that one model works well on different datasets, without having to optimize the penalties for each dataset.

\end{itemize}



\def\pz{\phantom{0}}

\begin{table}
    \centering
    \setlength{\tabcolsep}{4pt}
\renewcommand{\baselinestretch}{1.2}
    \small
    \begin{tabular}{|c|c|c|c|c|}
        \hline
        \multirow{2}{*}{Benchmarks} & \multirow{2}{*}{ACL} & arXiv & \multirow{2}{*}{Wiki+} & Wiki+ \\
        & & OCR & & no $\msp$ \\
        \hline
        Error distribution& 61-30-9 & 60-32-8 & 19-79-2 & 93-0-7 \\
        \hline
        \multicolumn{5}{c}{}\\[-3mm]
        \multicolumn{5}{c}{Percentage of corrected spelling and mixed errors}\\
        \hline

       Google           & 13.5\,\%  & 16.5\,\%  & 75.0\,\%  & \pz4.0\,\%  \\
       BID+Google       & 18.4\,\%  & 22.2\,\%  & 75.2\,\%  &   81.2\,\%  \\
       Oracle+Google    & 19.9\,\%  & 22.0\,\%  & 75.3\,\%  &   82.1\,\%  \\
        \hline
        \multicolumn{5}{c}{}\\[-3mm]
        \multicolumn{5}{c}{F-score for all errors combined}\\
        \hline

       Google           & 47.7\,\%  & 56.7\,\%  & 79.0\,\% & \pz8.2\,\%  \\
       BID+Google       & 66.6\,\%  & 76.3\,\%  & 83.4\,\% &   98.5\,\%  \\
       Oracle+Google    & 75.4\,\%  & 78.2\,\%  & 84.1\,\% &   98.9\,\%  \\
        \hline
    \end{tabular}
    \renewcommand{\baselinestretch}{1}
    \caption{
        Percentages of spelling errors corrected by Google only,
        Google after our tokenization repair (with BID+),
        and Google after perfect tokenization repair.
        The error distribution is given as three percentages: tokenization errors, spelling errors, mixed errors.
        }
    \label{tab:spelling}
\vspace{-0.5cm}
\end{table}


\begin{table}[ht]
 \renewcommand{\baselinestretch}{1.2}\small
 \begin{center}
 \begin{tabular}{|l|c|c|}
  \hline
  Approach      & F-score   & Seq.acc. \\
  \hline
  \citet{doval} & 99.6\,\%      & 92.2\,\%      \\
  UNI           & 99.6\,\%      & 90.7\,\%      \\
  BID           & {\bf 99.8}\,\%  & {\bf 94.2}\,\%  \\
  \hline
  \citet{nastase} & 40.3\,\%  & 19.0\,\%  \\
  BID+            & {\bf 89.6}\,\%  & {\bf 79.6}\,\%  \\
  \hline
 \end{tabular}
 \renewcommand{\baselinestretch}{1}
 \caption{Evaluation on the English benchmark by \citet{doval} and the ACL benchmark.
 }
    \label{tab:related}
 \end{center}
\vspace{-0.5cm}
\end{table}

Table \ref{tab:spelling} shows the capabilities of Google's error correction
(the best among all the existing tools, see Section \ref{sec:baselines})
with and without our tokenization repair on our four benchmarks with spelling mistakes.
We removed internal punctuation in tokens that were merged by our method or the oracle, to help Google to correct more errors.
The main takeaway is that Google can correct substantially more errors
when our tokenization repair is run as a pre-processing.
The reason is that once it is clear which characters form a word, 
it is much easier to figure out the correct word.
Furthermore, the upper bound by the Oracle+Google approach in Table \ref{tab:spelling} shows that using our method as a pre-processing reaches near optimal results for all benchmarks, except for the ACL benchmark which has severe errors.
The difference in corrected spelling and mixed errors between Google and Oracle+Google on the \textit{Wiki+} benchmark is small, because there are only very few mixed errors.

Note that there are \emph{two} reasons why the overall F-score of BID+Google is higher than for Google alone:
because our tokenization repair fixes many more tokenization errors than Google
and because our tokenization repair helps Google to fix more spelling errors.

\subsection{Evaluation of previous work}\label{sec:evaluation:previous}

The results for the best previous work are shown in Table \ref{tab:related}.
These works did not provide sufficient material to run their methods on our benchmarks.
Instead, we ran our approaches on their benchmarks.

The approach from \citet{doval} is an instance of UNI.
The BID model performs better than their model and our UNI, which shows the improvement by the bidirectional component.
Figures are high for all three models because their benchmark is unrealistic:
it has no spelling errors and all spaces are removed.
Our results in Table \ref{tab:eval-results} show that the problem is much harder with spelling errors
(in particular, see the low sequence accuracy for UNI in column \textit{Wiki+ no \msp}),
and that results are much worse when not making use of existing spaces
(compare the F-score and sequence accuracy for columns \textit{Wiki+} and \textit{Wiki+ no \msp}).

The results from \citet{nastase} on the ACL corpus are very weak in comparison.%
\footnote{\citet{nastase} report an F-score of $95\,\%$,
whereas Table~\ref{tab:related} states $40.3\,\%$.
There are three reasons for this:
(1) they remove all spaces and evaluate how many words they can restore, including words that had no error initially;
(2) they evaluate on newer documents which have less OCR errors;
(3) they make many errors around punctuation, but their word-level evaluation does not punish this.}

\subsection{Error analysis}\label{sec:evaluation:analysis}

Our interactive web application under \webpage\ allows a detailed error analysis
for all benchmarks and methods.
We here list our most important findings.

Methods trained without spelling errors have a strong tendency to split words with a typo because there is no meaningful continuation (e.g., \ourexample{unwnted pregnancies} is wrongly repaired to \ourexample{unw nted pregnancies}) and to wrongly merge misspelled words when they happen to form a correct word (e.g., if \ourexample{as well} is mistyped as \ourexample{s well}, it is wrongly repaired to \ourexample{swell}).%

The ACL benchmark has many passages that are hard to correct given only the input text.
In particular:
wrong reading order due to imperfect text extraction from the PDFs (e.g., \ourexample{Prepa- Seasoning ration}),
ambiguous formulas (e.g., does the text \ourexample{nik} come from \ourexample{$n_i \: k$} or \ourexample{$n_{ik}$}),
and very deformed words (e.g., \ourexample{(.hme:e~"s} with ground truth \ourexample{Chinese}).
The \emph{arXiv OCR} benchmark has, by construction, similar problems,
except the wrong reading order; hence the better figures in Table \ref{tab:eval-results}.

For the \emph{arXiv pdftotext} benchmark, many of the unresolved errors are in formulas
(often involving unusual mathematical symbols) and unconventionally written words
(e.g. \ourexample{bench mark}). Many of these could be considered as ground truth errors.

On Wikipedia, our method struggles mostly with compound words (e.g., \ourexample{offseason}),
entity names (Baseball team \ourexample{Waikiki BeachBoys} vs. rock band \ourexample{The Beach Boys}),
foreign words and inconsistent punctuation.
The few remaining errors were due to abbreviations, measuring units or rare symbols.

\section{Conclusion}\label{sec:conclusion}
We identified three key ingredients of high-quality tokenization repair, all missing from previous work:
bidirectional models, training on text with spelling errors, and making use of the space information already present. 
Our methods also improve existing spell checkers, by helping them to identify which characters form a word.

There is still room for improvement, especially in a scenario where many spelling (or OCR) errors and tokenization errors come together.
However, we show that the remaining errors are hard, with many ambiguous situations.
We paid attention to practical running times and all our methods have linear complexity.
A carefully trained Transformer model \cite{transformer}, however, has the potential to achieve comparable results with much faster running times \cite{sebastian-walter}.

It remains an open problem whether tokenization repair and spelling correction can be solved
``together'' both efficiently and with high quality.
Our own methods can be extended to also correct spelling errors,
but only with very large beam sizes and impractically large running times.
Existing models that correct both types of errors (and sometimes even grammar errors)
simultaneously fare poorly in comparison to the results we presented.

\section*{Acknowledgement}

We thank Claudius Korzen for providing the arXiv dataset, and Markus N\"ather for helpful discussions.

\bibliographystyle{acl_natbib}
\bibliography{tokenization-repair}

\begin{thebibliography}{34}
\expandafter\ifx\csname natexlab\endcsname\relax\def\natexlab#1{#1}\fi

\bibitem[{Abadi et~al.(2015)Abadi, Agarwal, Barham, Brevdo, Chen, Citro,
  Corrado, Davis, Dean, Devin, Ghemawat, Goodfellow, Harp, Irving, Isard, Jia,
  Jozefowicz, Kaiser, Kudlur, Levenberg, Man\'{e}, Monga, Moore, Murray, Olah,
  Schuster, Shlens, Steiner, Sutskever, Talwar, Tucker, Vanhoucke, Vasudevan,
  Vi\'{e}gas, Vinyals, Warden, Wattenberg, Wicke, Yu, and Zheng}]{tensorflow}
Mart\'{\i}n Abadi, Ashish Agarwal, Paul Barham, Eugene Brevdo, Zhifeng Chen,
  Craig Citro, Greg~S. Corrado, Andy Davis, Jeffrey Dean, Matthieu Devin,
  Sanjay Ghemawat, Ian Goodfellow, Andrew Harp, Geoffrey Irving, Michael Isard,
  Yangqing Jia, Rafal Jozefowicz, Lukasz Kaiser, Manjunath Kudlur, Josh
  Levenberg, Dandelion Man\'{e}, Rajat Monga, Sherry Moore, Derek Murray, Chris
  Olah, Mike Schuster, Jonathon Shlens, Benoit Steiner, Ilya Sutskever, Kunal
  Talwar, Paul Tucker, Vincent Vanhoucke, Vijay Vasudevan, Fernanda Vi\'{e}gas,
  Oriol Vinyals, Pete Warden, Martin Wattenberg, Martin Wicke, Yuan Yu, and
  Xiaoqiang Zheng. 2015.
\newblock \href {https://www.tensorflow.org/} {{TensorFlow: Large-Scale Machine
  Learning on Heterogeneous Systems}}.
\newblock Software available from tensorflow.org.

\bibitem[{Adesam et~al.(2019)Adesam, Dann{\'{e}}lls, and Tahmasebi}]{swedish}
Yvonne Adesam, Dana Dann{\'{e}}lls, and Nina Tahmasebi. 2019.
\newblock \href {http://ceur-ws.org/Vol-2364/1\_paper.pdf} {{Exploring the
  Quality of the Digital Historical Newspaper Archive KubHist}}.
\newblock In \emph{Proceedings of the Digital Humanities in the Nordic
  Countries 4th Conference, Copenhagen, Denmark}, {CEUR} Workshop Proceedings.

\bibitem[{Almuhareb et~al.(2019)Almuhareb, Alsanie, and
  Al{-}Thubaity}]{arabic-lstm}
Abdulrahman Almuhareb, Waleed Alsanie, and Abdulmohsen Al{-}Thubaity. 2019.
\newblock \href {https://doi.org/10.1109/ACCESS.2019.2893460} {Arabic word
  segmentation with long short-term memory neural networks and word embedding}.
\newblock \emph{{IEEE} Access}, 7:12879--12887.

\bibitem[{Attardi(2017)}]{attardi:wiki-extractor}
Giuseppe Attardi. 2017.
\newblock {WikiExtractor: A tool for extracting plain text from Wikipedia
  dumps}.
\newblock \url{https://github.com/attardi/wikiextractor}.

\bibitem[{Bast and Korzen(2017)}]{PDFs}
Hannah Bast and Claudius Korzen. 2017.
\newblock \href {https://doi.org/10.1109/JCDL.2017.7991564} {{A Benchmark and
  Evaluation for Text Extraction from PDF}}.
\newblock In \emph{{ACM/IEEE} Joint Conference on Digital Libraries, {JCDL},
  Toronto, ON, Canada}, pages 1--10. {IEEE} Computer Society.

\bibitem[{Bird et~al.(2008)Bird, Dale, Dorr, Gibson, Joseph, Kan, Lee, Powley,
  Radev, and Tan}]{acl-corpus}
Steven Bird, Robert Dale, Bonnie~J. Dorr, Bryan~R. Gibson, Mark~Thomas Joseph,
  Min{-}Yen Kan, Dongwon Lee, Brett Powley, Dragomir~R. Radev, and Yee~Fan Tan.
  2008.
\newblock \href
  {http://www.lrec-conf.org/proceedings/lrec2008/summaries/445.html} {The {ACL}
  anthology reference corpus: {A} reference dataset for bibliographic research
  in computational linguistics}.
\newblock In \emph{Proceedings of the International Conference on Language
  Resources and Evaluation, {LREC} 2008, 26 May - 1 June 2008, Marrakech,
  Morocco}. European Language Resources Association.

\bibitem[{Bird et~al.(2009)Bird, Klein, and Loper}]{nltk}
Steven Bird, Ewan Klein, and Edward Loper. 2009.
\newblock \href {http://www.oreilly.de/catalog/9780596516499/index.html}
  {\emph{Natural Language Processing with Python}}.
\newblock O'Reilly.

\bibitem[{Chiron et~al.(2017)Chiron, Doucet, Coustaty, and Moreux}]{icdar2017}
Guillaume Chiron, Antoine Doucet, Micka{\"{e}}l Coustaty, and Jean{-}Philippe
  Moreux. 2017.
\newblock \href {https://doi.org/10.1109/ICDAR.2017.232} {{ICDAR2017}
  competition on post-ocr text correction}.
\newblock In \emph{14th {IAPR} International Conference on Document Analysis
  and Recognition, {ICDAR} 2017, Kyoto, Japan, November 9-15, 2017}, pages
  1423--1428. {IEEE}.

\bibitem[{Devlin et~al.(2019)Devlin, Chang, Lee, and Toutanova}]{bert}
Jacob Devlin, Ming{-}Wei Chang, Kenton Lee, and Kristina Toutanova. 2019.
\newblock \href {https://doi.org/10.18653/v1/n19-1423} {{BERT:} pre-training of
  deep bidirectional transformers for language understanding}.
\newblock In \emph{Proceedings of the 2019 Conference of the North American
  Chapter of the Association for Computational Linguistics: Human Language
  Technologies, {NAACL-HLT} 2019, Minneapolis, MN, USA, June 2-7, 2019, Volume
  1 (Long and Short Papers)}, pages 4171--4186. Association for Computational
  Linguistics.

\bibitem[{Dong and Smith(2018)}]{ocr-dong}
Rui Dong and David Smith. 2018.
\newblock \href {https://doi.org/10.18653/v1/P18-1220} {Multi-input attention
  for unsupervised {OCR} correction}.
\newblock In \emph{Proceedings of the 56th Annual Meeting of the Association
  for Computational Linguistics, {ACL} 2018, Melbourne, Australia, July 15-20,
  2018, Volume 1: Long Papers}, pages 2363--2372. Association for Computational
  Linguistics.

\bibitem[{Doval and G{\'{o}}mez{-}Rodr{\'{\i}}guez(2019)}]{doval}
Yerai Doval and Carlos G{\'{o}}mez{-}Rodr{\'{\i}}guez. 2019.
\newblock \href {https://doi.org/10.1002/asi.24082} {{Comparing neural- and
  N-gram-based language models for word segmentation}}.
\newblock \emph{Journal of the Association for Information Science and
  Technology}, 70:187--197.

\bibitem[{FooLabs(2014)}]{pdftotext}
FooLabs. 2014.
\newblock \href {http://www.foolabs.com/xpdf} {{Xpdf: A PDF Viewer for X}}.

\bibitem[{Graves(2013)}]{graves:generating}
Alex Graves. 2013.
\newblock \href {http://arxiv.org/abs/1308.0850} {{Generating Sequences With
  Recurrent Neural Networks}}.
\newblock \emph{CoRR}, abs/1308.0850.

\bibitem[{Hagiwara and Mita(2020)}]{github-corpus}
Masato Hagiwara and Masato Mita. 2020.
\newblock \href {https://www.aclweb.org/anthology/2020.lrec-1.835} {{G}it{H}ub
  typo corpus: A large-scale multilingual dataset of misspellings and
  grammatical errors}.
\newblock In \emph{Proceedings of the 12th Language Resources and Evaluation
  Conference}, pages 6761--6768, Marseille, France. European Language Resources
  Association.

\bibitem[{Halpern(2015)}]{beautifuldata}
Orit Halpern. 2015.
\newblock \emph{{Beautiful data: A history of vision and reason since 1945}}.
\newblock Duke University Press.

\bibitem[{H{\"{a}}m{\"{a}}l{\"{a}}inen and Hengchen(2019)}]{helsinki}
Mika H{\"{a}}m{\"{a}}l{\"{a}}inen and Simon Hengchen. 2019.
\newblock \href {https://doi.org/10.26615/978-954-452-056-4\_051} {{From the
  Paft to the Fiiture: a Fully Automatic {NMT} and Word Embeddings Method for
  OCR Post-Correction}}.
\newblock In \emph{Proceedings of the International Conference on Recent
  Advances in Natural Language Processing, {RANLP} 2019, Varna, Bulgaria}.
  {INCOMA} Ltd.

\bibitem[{Huang et~al.(2020)Huang, Cheng, Chen, Wang, and Chu}]{chinese-bert}
Weipeng Huang, Xingyi Cheng, Kunlong Chen, Taifeng Wang, and Wei Chu. 2020.
\newblock \href {https://doi.org/10.18653/v1/2020.coling-main.186} {Towards
  fast and accurate neural chinese word segmentation with multi-criteria
  learning}.
\newblock In \emph{Proceedings of the 28th International Conference on
  Computational Linguistics, {COLING} 2020, Barcelona, Spain (Online), December
  8-13, 2020}, pages 2062--2072. International Committee on Computational
  Linguistics.

\bibitem[{Kingma and Ba(2015)}]{adam}
Diederik~P. Kingma and Jimmy Ba. 2015.
\newblock \href {http://arxiv.org/abs/1412.6980} {{Adam: A Method for
  Stochastic Optimization}}.
\newblock In \emph{3rd International Conference on Learning Representations,
  {ICLR} , San Diego, CA, USA, Conference Track Proceedings}.

\bibitem[{Kissos and Dershowitz(2016)}]{ocr-kissos}
Ido Kissos and Nachum Dershowitz. 2016.
\newblock \href {https://doi.org/10.1109/DAS.2016.44} {{OCR} error correction
  using character correction and feature-based word classification}.
\newblock In \emph{12th {IAPR} Workshop on Document Analysis Systems, {DAS}
  2016, Santorini, Greece, April 11-14, 2016}, pages 198--203. {IEEE} Computer
  Society.

\bibitem[{Kukich(1992)}]{kukich}
Karen Kukich. 1992.
\newblock \href {https://doi.org/10.1145/129875.129882} {{Spelling Correction
  for Telecommunications Network for the Deaf}}.
\newblock \emph{Communications of the {ACM}}, 35:80--90.

\bibitem[{Ma et~al.(2018)Ma, Ganchev, and Weiss}]{chinese-lstm}
Ji~Ma, Kuzman Ganchev, and David Weiss. 2018.
\newblock \href {https://doi.org/10.18653/v1/d18-1529} {State-of-the-art
  chinese word segmentation with bi-{LSTM}s}.
\newblock In \emph{Proceedings of the 2018 Conference on Empirical Methods in
  Natural Language Processing, Brussels, Belgium, October 31 - November 4,
  2018}, pages 4902--4908. Association for Computational Linguistics.

\bibitem[{Medress et~al.(1977)Medress, Cooper, Forgie, Green, Klatt, O'Malley,
  Neuburg, Newell, Reddy, Ritea, Shoup{-}Hummel, Walker, and
  Woods}]{beam-search}
Mark~F. Medress, Franklin~S. Cooper, James~W. Forgie, C.~C. Green, Dennis~H.
  Klatt, Michael~H. O'Malley, Edward~P. Neuburg, Allen Newell, Raj Reddy,
  H.~Barry Ritea, J.~E. Shoup{-}Hummel, Donald~E. Walker, and William~A. Woods.
  1977.
\newblock \href {https://doi.org/10.1016/0004-3702(77)90026-1} {Speech
  understanding systems}.
\newblock \emph{Artif. Intell.}, 9(3):307--316.

\bibitem[{Mik{\v{s}}a et~al.(2010)Mik{\v{s}}a, {\v{S}}najder, and
  Ba{\v{s}}ic}]{croatian}
Mladen Mik{\v{s}}a, Jan {\v{S}}najder, and Bojana~Dalbelo Ba{\v{s}}ic. 2010.
\newblock \href
  {http://dcl.bas.bg/wp-content/uploads/2015/08/FASSBL7_2010_proceedings-3.pdf#page=68}
  {{Correcting Word Merge Errors in Croatian Texts}}.
\newblock \emph{The Seventh International Conference on Formal Approaches to
  South Slavic and Balkan Languages}.

\bibitem[{Nastase and Hitschler(2018)}]{nastase}
Vivi Nastase and Julian Hitschler. 2018.
\newblock \href
  {http://www.lrec-conf.org/proceedings/lrec2018/summaries/114.html}
  {{Correction of OCR Word Segmentation Errors in Articles from the ACL
  Collection through Neural Machine Translation Methods}}.
\newblock In \emph{Proceedings of the Eleventh International Conference on
  Language Resources and Evaluation, {LREC}, Miyazaki, Japan}. European
  Language Resources Association {(ELRA)}.

\bibitem[{Nguyen et~al.(2019)Nguyen, Jatowt, Coustaty, Nguyen, and
  Doucet}]{ocr-nguyen19}
Thi{-}Tuyet{-}Hai Nguyen, Adam Jatowt, Micka{\"{e}}l Coustaty, Nhu{-}Van
  Nguyen, and Antoine Doucet. 2019.
\newblock \href {https://doi.org/10.1109/ICDAR.2019.00145} {Post-{OCR} error
  detection by generating plausible candidates}.
\newblock In \emph{2019 International Conference on Document Analysis and
  Recognition, {ICDAR} 2019, Sydney, Australia, September 20-25, 2019}, pages
  876--881. {IEEE}.

\bibitem[{Nguyen et~al.(2020)Nguyen, Jatowt, Nguyen, Coustaty, and
  Doucet}]{ocr-nguyen}
Thi{-}Tuyet{-}Hai Nguyen, Adam Jatowt, Nhu{-}Van Nguyen, Micka{\"{e}}l
  Coustaty, and Antoine Doucet. 2020.
\newblock \href {https://doi.org/10.1145/3383583.3398605} {Neural machine
  translation with {BERT} for post-{OCR} error detection and correction}.
\newblock In \emph{{JCDL} '20: Proceedings of the {ACM/IEEE} Joint Conference
  on Digital Libraries in 2020, Virtual Event, China, August 1-5, 2020}, pages
  333--336. {ACM}.

\bibitem[{Niklas(2010)}]{ocr-niklas}
Kai Niklas. 2010.
\newblock \href {http://www.l3s.de/~tahmasebi/Diplomarbeit_Niklas.pdf}
  {Unsupervised post-correction of {OCR} errors}.
\newblock \emph{Diploma thesis. Leibniz Universit{\"a}t Hannover}.

\bibitem[{Rigaud et~al.(2019)Rigaud, Doucet, Coustaty, and Moreux}]{icdar2019}
Christophe Rigaud, Antoine Doucet, Micka{\"{e}}l Coustaty, and Jean{-}Philippe
  Moreux. 2019.
\newblock \href {https://doi.org/10.1109/ICDAR.2019.00255} {{ICDAR} 2019
  competition on post-{OCR} text correction}.
\newblock In \emph{2019 International Conference on Document Analysis and
  Recognition, {ICDAR} 2019, Sydney, Australia, September 20-25, 2019}, pages
  1588--1593. {IEEE}.

\bibitem[{Soni et~al.(2019)Soni, Klein, and Eisenstein}]{historical}
Sandeep Soni, Lauren~F. Klein, and Jacob Eisenstein. 2019.
\newblock \href {https://doi.org/10.18653/v1/w19-2513} {{Correcting Whitespace
  Errors in Digitized Historical Texts}}.
\newblock In \emph{Proceedings of the 3rd Joint {SIGHUM} Workshop on
  Computational Linguistics for Cultural Heritage, Social Sciences, Humanities
  and Literature, LaTeCH@NAACL-HLT, Minneapolis, MN, USA}, pages 98--103.
  Association for Computational Linguistics.

\bibitem[{Taghva and Stofsky(2001)}]{ocr-spell}
Kazem Taghva and Eric Stofsky. 2001.
\newblock \href {https://doi.org/10.1007/PL00013558} {{OCRS}pell: an
  interactive spelling correction system for {OCR} errors in text}.
\newblock \emph{Int. J. Document Anal. Recognit.}, 3(3):125--137.

\bibitem[{Tong and Evans(1996)}]{ocr-tong}
Xiang Tong and David~A. Evans. 1996.
\newblock \href {https://www.aclweb.org/anthology/W96-0108/} {A statistical
  approach to automatic {OCR} error correction in context}.
\newblock In \emph{Fourth Workshop on Very Large Corpora, VLC@COLING 1996,
  Copenhagen, Denmark, August 4, 1996}.

\bibitem[{Vaswani et~al.(2017)Vaswani, Shazeer, Parmar, Uszkoreit, Jones,
  Gomez, Kaiser, and Polosukhin}]{transformer}
Ashish Vaswani, Noam Shazeer, Niki Parmar, Jakob Uszkoreit, Llion Jones,
  Aidan~N. Gomez, Lukasz Kaiser, and Illia Polosukhin. 2017.
\newblock \href
  {https://proceedings.neurips.cc/paper/2017/hash/3f5ee243547dee91fbd053c1c4a845aa-Abstract.html}
  {Attention is all you need}.
\newblock In \emph{Advances in Neural Information Processing Systems 30: Annual
  Conference on Neural Information Processing Systems 2017, December 4-9, 2017,
  Long Beach, CA, {USA}}, pages 5998--6008.

\bibitem[{Viterbi(1967)}]{viterbi}
Andrew~J. Viterbi. 1967.
\newblock \href {https://doi.org/10.1109/TIT.1967.1054010} {{Error bounds for
  convolutional codes and an asymptotically optimum decoding algorithm}}.
\newblock \emph{{IEEE} Transactions on Information Theory}, 13:260--269.

\bibitem[{Walter(2021)}]{sebastian-walter}
Sebastian Walter. 2021.
\newblock Tokenization repair using transformers.
\newblock
  \url{https://ad-blog.cs.uni-freiburg.de/post/tokenization-repair-using-transformers}.
\newblock Accessed: 2021-09-07.

\end{thebibliography}
\end{document}